# IWA: Integrated Gradient based White-box Attacks for Fooling Deep Neural Networks

Yixiang Wang, Jiqiang Liu, Xiaolin Chang, Jelena Mišić, and Vojislav B. Mišić
*Abstract*—The widespread application of deep neural network (DNN) techniques is being challenged by adversarial examples—the legitimate input added with imperceptible and well-designed perturbations that can fool DNNs easily in the DNN testing/deploying stage. Previous adversarial example generation algorithms for adversarial white-box attacks used Jacobian gradient information to add perturbations. This information is too imprecise and inexplicit, which will cause unnecessary perturbations when generating adversarial examples.

This paper aims to address this issue. We first propose to apply a more informative and distilled gradient information, namely integrated gradient, to generate adversarial examples. To further make the perturbations more imperceptible, we propose to employ the restriction combination of $L_0$ and $L_1/L_2$ secondly, which can restrict the total perturbations and perturbation points simultaneously. Meanwhile, to address the non-differentiable problem of $L_1$, we explore a proximal operation of $L_1$ thirdly. Based on these three works, we propose two *I*ntegrated gradient based *W*hite-box *A*dversarial example generation algorithms (IWA): IFPA and IUA. IFPA is suitable for situations where there are a determined number of points to be perturbed. IUA is suitable for situations where no perturbation point number is preset in order to obtain more adversarial examples. We verify the effectiveness of the proposed algorithms on both structured and unstructured datasets, and we compare them with five baseline generation algorithms. The results show that our proposed algorithms do craft adversarial examples with more imperceptible perturbations and satisfactory crafting rate. $L_2$ restriction is more suitable for unstructured dataset and $L_1$ restriction performs better in structured dataset.

## I. Introduction

The fast rise of artificial intelligence in the past few years stems from the advancement of deep learning (DL) techniques, and DL has made fantastic success and state-of-the-art performance in various fields [1][2][3]. However, adversarial examples [4] challenge the security and reliability of DL techniques and then affect the widespread application of DL in security-critical settings. Adversarial examples are the legitimate input added with tiny and well-designed perturbation, which can cause deep neural networks (DNNs) misclassification and then cause the sharp degradation of their performance. Usually, the perturbation is hard, if not impossible, to be identified by human eyes. Since then, considerable researchers have devoted efforts to the study of adversarial examples mainly from two aspects: attacks and defenses [5]. The past years witnessed diverse adversarial example generation algorithms with more imperceptible perturbation and more aggressive skills, which can be applied to conduct white-box and/or black-box attacks [6][7][8][9]. These algorithms were also investigated in various fields such as sentiment analysis [7], malware detection [10], and automatic speech recognition [11] and many more. Meanwhile, defensive techniques to detect and/or resist adversarial examples have been proposed, such as gradient masking and obfuscated gradients [12].

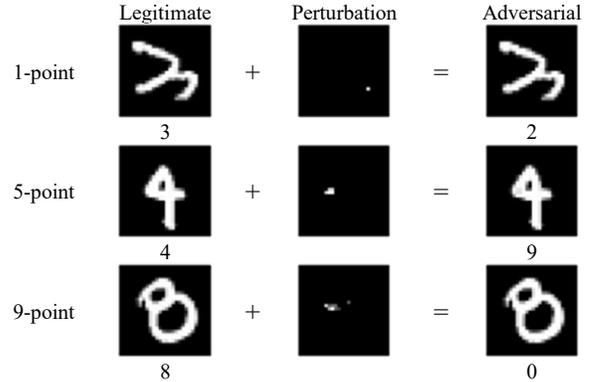

Fig.1. The adversarial examples generated by our proposed IFPA algorithm with specific points on MNIST dataset. Legitimate denotes the legitimate input and adversarial means the corresponding adversarial example. The number under each image represents the category the model thinks.

In this paper, we focus on adversarial example generation algorithms for adversarial white-box attacks. There existed two primary problems with previous white-box adversarial works as follows:

1) They used Jacobian gradient information as the basis for the addition of perturbations, which will cause unnecessary perturbations. Concretely, in the adversarial white-box attacks, the information provided by the Jacobian gradient is not clear and does not reflect what a DNN model learns [14]. This can cause redundant and undependable perturbations when this information is used to generate adversarial examples.

2) They were designed for a specific type of datasets, and their capability of generating high-quality adversarial examples was only verified on image unstructured datasets [15][16][17]. For instance, Carlini & Wanger (CW) attack [8] verified their algorithms only on image

Yixiang Wang, Jiqiang Liu (Corresponding author) and Xiaolin Chang are with Beijing Key Laboratory of Security and Privacy in Intelligent Transportation, Beijing Jiaotong University, Beijing 100044, China (e-mail: {18112047, jqliu, xlchang}@bjtu.edu.cn).

Jelena Mišić and Vojislav B. Mišić are with Ryerson University, Toronto, Canada (e-mail: {jmisic,vmisic}@ryerson.ca ).

unstructured datasets such as MNIST, CIFAR-10 and ImageNet. Grosse *et al.* [18] verified their attack only on a structured malware dataset. Neither of them considered the applicability of the proposed algorithms on the other type datasets. We think it is not comprehensive to verify the algorithm on the single type datasets. Here, by structured data, we mean it resembles one table in a database. Unstructured data is like video, image, etc. [13].

In this paper, we propose two *I*ntegrated gradient based *W*hite-box *A*dversarial example generation algorithms (IWA): Integrated-based Finite Point Attack algorithm (IFPA) and Integrated-based Universe Attack algorithm (IUA), to address these concerns. IFPA is suitable for situations where there are a determined number of points to be perturbed. IUA is suitable for situations where no perturbation point number is preset in order to obtain more adversarial examples. Both algorithms have the following three major features:

1) Use the integrated gradient to generate adversarial examples. Compared with the Jacobian gradient, the integrated gradient contains more informative and distilled gradient information and can better reflect each input point's contribution to the DNN output. It can guide to select points to be perturbed. In the structured dataset, one point represents one actual value in an item. In the image unstructured dataset, differs from one pixel whose definition is three-channel values in a certain position of an RGB image, one point is only one channel value.

2) Employ $L_p$ norm restriction to limit the perturbations added on each point. Previous work has proven the effectiveness of $L_p$ norm restriction [8]. Both two algorithms achieve the combination $L_0$ with $L_1/L_2$ restriction. That is, when generating adversarial examples, the proposed algorithms not only restrict the perturbation points, but also restrict the total perturbations.

3) Employ a proximal operation of $L_1$ to address the non-differentiable problem of $L_1$. $L_1$ restriction rarely appears in adversarial generation algorithms since it is not differentiable [19]. Carlini *et al.* [20] tried to use an algorithm to approximate the $L_1$ restriction. In this paper, we apply the proximal operation [21] mathematically to solve it.

We verify the effectiveness of the proposed two algorithms on the structured dataset, NSL-KDD, and unstructured datasets, MNIST and CIFAR-10. The experimental results verify that our algorithms can generate adversarial examples with more imperceptible perturbations and satisfactory crafting rate compared with the prior algorithms. Also, results indicate that $L_2$ restriction is more suitable for unstructured datasets and $L_1$ restriction performs better in structured datasets. We hope that our proposed attack can provide guidelines to researches on adversarial example defense.

The rest of the paper is organized as follows. In Section II, we present the notations, and describe the background and development of adversarial examples. Problem definition and the proposed two algorithms are provided in Section III. Section IV is the evaluation of extensive experimental results on comparative algorithms and ours. Conclusion and future work are investigated in Section V.

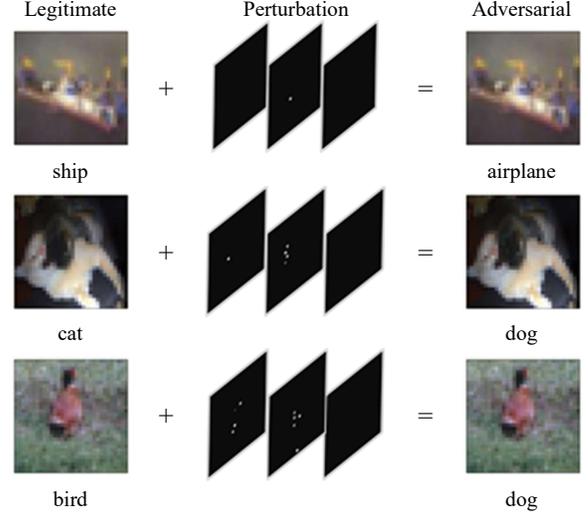

Fig.2  Adversarial examples generated by IFPA algorithm with 1-point, 5-point and 9-point attack on CIFAR-10 dataset.

## II. BACKGROUND AND RELATED WORK

This section first presents the notations and background knowledge of adversarial examples. Then, we present the related work and five comparable baseline algorithms.

*A. Background*

*1) Neural Network and Notations*

Given the legitimate input $x \in \mathbb{R}^n$, DNNs can output $y \in \mathbb{R}^m$, and can be formalized as a sophisticated function with multiple layers:

$$y = f_k\left(f_{k-1}\left(...f_2\left(f_1(x;\theta_1);\theta_2\right)...\right);\theta_k\right) = F(x;\theta)$$

where $k$ is the $k^{th}$ layer, and $f_k(x;\theta_k)$ denotes what the function of the $k^{th}$ layer learns with parameters $\theta_k$. Each layer function $f$ is unified as:

$$f(x;\theta) = \sigma\left(\theta_{weight} \cdot x\right) + \theta_{bias}$$

where $\sigma(\cdot)$ is a non-linear activation function such as *ReLU* [22], *LeakyReLU* [23] and *softmax*. $\theta_{weight}$ and $\theta_{bias}$ in each layer combine the model parameters $\theta$. In this paper, we consider the model as an *m*-class classifier. The output of the model is transformed to the probability distribution of *m* classes via *softmax* function in the last layer. That is, we refer $F_i(x)$ to the probability of the $i^{th}$ class and every element of $F(x)$ satisfies: i) $\sum_{i=1}^{m} F(x)_i = 1$ and ii) $\forall F(x), F(x) \in [0,1]$. Then, we believe $argmax(F(x))$ as the predicted label of $x$, $C(x)$, and the ground truth of $x$ is $C^*(x)$. So, another formalization of $F$ is:

$$F(x;\theta) = softmax(Z(x;\theta)) = y \qquad (1.1)$$

$F(x;\theta)$ is the complete function a DNN model learns, and $F(x;\theta)$ includes the *softmax* activation function. $Z(x;\theta)$ is the output without *softmax* function and is called this logit output. In this paper, we pay attention to generating adversarial examples on the logit output.

*2) Adversarial Example and Thread Model*

Szegedy *et al.* [4] was the first to uncover the adversarial examples: given the legitimate input $x$ and its ground truth $C^*(x)$, the adversary can find $x_{adv}/x'$ ($x_{adv}$ and $x'$ denote adversarial examples in this paper) that is very similar to $x$ but $C(x_{adv}) \neq C^*(x)$. And $C(x_{adv})$ produces different results according to the adversary's knowledge and goals.

When it comes to attacks, it is necessary to quantify the enemy's capabilities. Papernot *et al.* [24] firstly decomposed the space of adversaries in the deep learning systems, according to the adversarial goals and knowledge. In this paper, we follow their work [5][24] to classify all adversarial attacks. That is, according to the adversary's knowledge, we divide the attack into a black box attack and white box attack:

- White-box attacks assume that the adversary has access to a neural network, which means that the adversary knows the model topology, model parameters and the data training for the model.
- Black-box attacks assume that the adversary has no access to a neural network, which means that the adversary has no idea of the model topology and parameters, but knows the model output.

According to the adversary's goals, we can categorize the attacks into the targeted attacks and non-target attacks:

- Non-target attacks are attacks that adversarial example generated by adversaries cause the model to output the arbitrary class except the ground truth $C^*(x)$, i.e., $C(x_{adv}) \neq C^*(x)$.
- Targeted attacks mean that the adversarial example $x_{adv}$ generated by the adversary can cause the model to output the class $t$ that is specified by the adversary, where this class differs from the ground truth class $C^*(x)$, i.e., $t = C(x_{adv}) \neq C^*(x)$.

In this paper, we assume the adversary has the ability to carry out white-box attacks, and his goal is to cause a non-target attack. It is reasonable since the previous work has shown that the nature of adversarial example, namely transferability, can cause model $B$ vulnerable to adversarial examples generated by model $A$ [25]. So, if the adversary implements a black-box attack on model $B$, it can attack the self-defined model $A$ in a white-attack manner. This is equivalent to a white-box attack to some extent.

*3) Integrated Gradient*

Sundararajan *et al.* [14] proposed a method to solve the problem of attributing the prediction of a deep network to its input features. That is, the proposed method, integrated gradient, can calculate the contribution of each input element to the specific class. The formula of the integrated gradient is as follows.

$$IntegratedGradient(x) = \left(x_i - x_i^{baseline}\right) \times \sum_{j=1}^{S} \frac{\partial Z\left(x^{baseline} + \frac{j}{S} \times \left(x - x^{baseline}\right)\right)}{\partial x_i} \times \frac{1}{S} \quad (II.1)$$

where $x^{baseline}$ is the baseline example in their settings, for instance, the pure black image is a baseline example in image data; all zero-embedding vector is a baseline in text data. $S$ is the iteration steps. As shown in Fig. 3, we give an example to see the difference between the Jacobian gradients of $Z(x)$ w.r.t. $x$ and the integrated gradients. We can see that integrated gradients are more precise than jacobian gradients and have more detailed textures at the object's edge. Thus, we think using the integrated gradient can generate adversarial examples better than the Jacobian gradient.

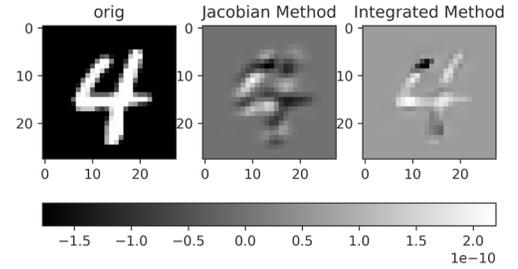

Fig. 3. The comparison between the Jacobian gradient and the integrated gradient on the MNIST sample '4'.

*4) $L_p$-norm Distance*

$L_p$-norm distance is an evaluation metric to assess the gap between two vectors in a vector space. We need this to evaluate the difference between adversarial examples and legitimate examples. The definition of $L_p$-norm of $x$ and $x'$ is shown below.

$$L_p(x - x') = \left(\sum_{i=1}^{n} |x_i - x'_i|^p\right)^{1/p} \quad (II.2)$$

Specifically:
- $L_0$ measures the number of different points in $x$ and $x'$, i.e. $\sum_{i=1}^{n} I(x_i \neq x'_i)$, where $I(\cdot)$ is the indicator function that when the inside condition is true, $I(\cdot) = 1$.
- $L_1$ measures the sum of the absolute values of differences at all points, i.e., $\sum_{i=1}^{n} |x_i - x'_i|$.
- $L_2$ is the well-known Euclidean distance, whose formula is $\sqrt{\sum_{i=1}^{n} (x_i - x'_i)^2}$.

In our experiments, these three $L_p$ distances will be discussed, and $L_0$ will be combined with $L_1$ or $L_2$ to restrict the perturbations between $x$ and $x'$, which will be described in Section III. In machine learning algorithms, $L_p$-norm is also called regularizer $L_p$.

## B. Related Work

As mentioned above, Szegedy *et al.* [4] applied L-BFGS algorithm to uncover the existence of adversarial examples. Then, FGSM [15] was proposed to use the gradient of the model to generate adversarial examples. Many follow-up algorithms were affected by it to generate adversarial examples via gradient information. Compared with previous works, instead of using Jacobian gradient information to generate adversarial examples, we apply a sophisticated gradient, the integrated gradient, to obtain the informative gradient.

Certainly, some algorithms do not use gradient information to generate adversarial examples such as one-pixel [17], DeepFool [26], Nattack [27]. For example, zeroth order optimization (ZOO) [28] attack uses approximate gradients by a finite difference method. This means that the adversary can still generate adversarial examples without obtaining the model information. In this paper, we only consider white-box attacks on the model gradients.

As for the perturbation restriction, adversarial generation algorithms used $L_p$ to constrain the perturbations. For instance, Deepfool [26] places $L_2$ restriction in the divisor part. CW [8] used $L_0, L_2, L_\infty$ to restrict the total perturbations, respectively. But there is no research on how to utilize $L_1$ fully since $L_1$ is not differentiable in the vector space. Previously, there was only one work to deal with $L_1$ restriction [20]. But they used the Reluplex algorithm to achieve similar $L_0$ restriction. Differs from their work, we introduce the proximal $L_1$ to solve $L_1$ non-differentiable problem, from the perspective of mathematical theory. Also, how to combine $L_0$ and $L_1/L_2$ is necessary. One-pixel attack achieves the goal of restricting $L_0$, but it did not use $L_1$ or $L_2$ to restrict the total perturbations so that the perturbation points can be found clearly. Moreover, the core idea of the one-pixel attack is the differential evolution algorithm, not the gradient information. In this paper, our proposed IFPA algorithm solves the problem mentioned above.

Mechanisms to defend against adversarial examples were also proposed by researchers. Distillation was proposed by Papernot *et al.* [29] to defend DNN against adversarial examples. Concretely, they divided the logit output by temperature $T$ to control the degree of distillation. But the emergency of CW attack proved that distillation cannot resist CW attack. Afterward, other work [30] proved some defensive algorithms were not as robust as they declared. In this paper, we do not pay attention to how to break through the defensive algorithms, but focus on how to generate high-quality adversarial examples. From the perspective of experimental results, our proposed algorithms can generate adversarial examples better than that of previous algorithms described in the next section.

## C. Adversarial Example Generation Algorithms

Thus far, several studies have proposed classical white-box adversarial example generation algorithms. Here, in order to facilitate the comparison between our algorithms and these algorithms, we briefly present each algorithm.

### 1) FGSM [15]

Fast Gradient Sign Method algorithm was proposed to generate adversarial examples by:

$$x_{adv} = x + \varepsilon sign(\nabla_x J(x, C^*(x)))$$

where $sign(v)$ denotes the sign of every element in the vector $v$, and $\nabla_x J(\cdot)$ denotes the derivate of the loss function $J(\cdot)$ w.r.t. $x$. $\varepsilon$ is the one-step perturbation ratio. This is a simple way to craft $x_{adv}$. FGSM inspires a lot of subsequent work, including BIM.

### 2) BIM [16]

BIM is an iterative multi-step variant of FGSM. The iterative function is defined as:

$$x'_{N+1} = Clip(x'_N + \varepsilon sign(\nabla_x J(x, C^*(x))), x - \epsilon, x + \epsilon)$$

where function $Clip(v, lb, ub)$ restricts every element in $v$ between the lower bound $lb$ and the upper bound $ub$. $N$ is the iterations. $\epsilon$ is the total perturbations. BIM solves the problem of lower adversarial example crafting rate in FGSM.

### 3) DeepFool [26]

DeepFool aims to find the minimum distance between the sample and its nearest decision boundary via projection. Two critical components of DeepFool is: 1) find the nearest decision boundary $\hat{l}(x)$ w.r.t. $x_0$; and 2) calculate the distance to the decision boundary. The distance is what we call perturbation. The formalization description is shown below.

$$\hat{l}(x_0) = \arg\min_{C \neq C^*} \frac{\left|F_C(x_0) - F_{C^*(x_0)}(x_0)\right|}{\left|\theta_C - \theta_{C^*(x_0)}\right|_2}$$

$$dist(x_0) = \frac{\left|F_{\hat{l}(x_0)}(x_0) - F_{C^*(x_0)}(x_0)\right|}{\left|\theta_{\hat{l}(x_0)} - \theta_{C^*(x_0)}\right|_2^2} \left(\theta_{\hat{l}(x_0)} - \theta_{C^*(x_0)}\right)$$

### 4) CW attack [8]

Carlini & Wagner (CW) attack is proposed to verify that the defensive distillation method [29] is invalid for some attacks. CW attack can easily break the distilled model. The objective function of CW attack is that:

$$\text{minimize } L_p(\epsilon) + c \cdot g(x + \epsilon)$$
$$s.t. \quad x + \epsilon \in [0,1]^n$$

where $g(\cdot)$ is self-designed function:

$$g(x + \epsilon) = \max\left(\max\left(Z(x+\epsilon)_{C^*(x)}\right) - Z(x+\epsilon)_t, -\kappa\right)$$

The objective function differs from ours described in Eq.(III.3). CW attack optimizes the $L_p$ perturbations directly, but we use the $L_p$ distance between $x$ and $x'$. These two algorithms are different when implementing them. Also, the penalty term is different, CW attack penalizes their function $g(\cdot)$, we penalize the $L_p$ distance. Our objective function is mainly to minimize $Z_{C^*(x)}(x_{adv})$, but theirs is to minimize the total perturbation.

*5) PGD [9]*

Projected Gradient Descent (PGD) attack is another multi-step variant of FGSM. The formula is as follows:

$$x'_{N+1} = \prod\left(x'_N + \varepsilon sign\left(\nabla_x J\left(x, C^*(x)\right)\right)\right)$$

More commonly, adversarial examples generated by PGD are used to do adversarial training, a method to make the model more robust. But in essence, this is still an algorithm of adversarial example generation.

Table I is the comparison between previous works and our work. 'Algo.' means the comparable algorithms. 'Dataset' indicates the type of experimental dataset when the algorithm is proposed. The adversarial goal has been explained in Section II.A.2). 'Perturbation restriction' means that the algorithm uses the $L_p$ norm to restrict the perturbation or not. 'Attack frequency' represents the algorithm to generate adversarial examples in an iterative way or one-step way. These five adversarial generation algorithms are compared with ours, and our method's effectiveness is verified in Section IV.

TABLE I. COMPARISON OF RELATED WORKS FOR ADVERSARIAL GENERATION ALGORITHM

| Algo. | Dataset | Perturbation Restriction | | | Attack Frequency |
|---|---|---|---|---|---|
| | | $L_0$ | $L_1$ | $L_2$ | |
| FGSM | unstructured | - | - | - | one-step |
| BIM | unstructured | - | - | - | iterative-step |
| DeepFool | unstructured | - | - | ✓ | iterative-step |
| CW | unstructured | ✓ | - | ✓ | iterative-step |
| PGD | unstructured | - | - | ✓ | iterative-step |
| Ours | structured & unstructured | ✓ | ✓ | ✓ | iterative-step |

III. METHODOLOGY

In this section, we introduce the definition of the problem of adversarial examples in Section III.A, which will be the fundamental basis to compose generation algorithms. Then, we propose two generation algorithms and explain them in detail. Finally, we outline the evaluation metrics used in the next experimental section.

*A. Problem Definition*

Previously, there were two common optimization ways to generate adversarial examples. The first is maximizing the loss value of $x_{adv}$ in the direction of $C^*(x)$. At the same time, $x_{adv}$ should satisfies: 1) each element of $x_{adv}$ should be legitimate, i.e., $x_{adv} \in [0,1]^n$; and 2) the difference between $x_{adv}$ and legitimate $x$ should be smaller than $\epsilon$, where $\epsilon$ is as smaller as possible. In this way, the objective function with constraints can be formalized in Eq.(III.1).

$$\begin{aligned} maxmize \quad & J\left(F(x_{adv}), C^*(x)\right) \\ s.t. \quad & x_{adv} \in [0,1]^n \\ & |x - x_{adv}| < \epsilon \end{aligned} \quad (\text{III}.1)$$

The second is minimizing the difference between $x_{adv}$ and legitimate $x$ while satisfies $x_{adv} \in [0,1]^n$, and the model output of $x_{adv}$ should be as far as possible from that of $x$. The formula is shown in Eq. (III.2).

$$\begin{aligned} minimize \quad & |x - x_{adv}| \\ s.t. \quad & x_{adv} \in [0,1]^n \\ & F(x) \neq F(x_{adv}) \end{aligned} \quad (\text{III}.2)$$

In this paper, we combine objective functions and constraints in Eq. (III.1) and Eq. (III.2) and transform them to form a new objective function, shown in Eq. (III.3).

$$\begin{aligned} minimize \quad & Z_{C^*(x)}(x_{adv}) + c \cdot L_p(x, x_{adv}) \\ s.t. \quad & x_{adv} \in [0,1]^n \end{aligned} \quad (\text{III}.3)$$

That is, we aim to minimize two items at the same time. The first item is minimizing the logit output w.r.t. $x_{adv}$ at the ground truth $C^*(x)$. The second item is to minimize the gap between $x$ and $x_{adv}$ using $L_p$ distance described in Section II.D. Here, $c$ is a positive constant that is adopted to control the proportion of the distance part in the objective function. We refer to $c \cdot L_p(x, x_{adv})$ as $R_p$ for the convenience of the experimental part. However, when we specify $L_p$ as $L_1$, there is one issue: $L_1$ is not differentiable in the whole vector space [19]. To solve this, we propose to use the proximal operation of $L_1$ [21]:

$$Prox_{L_1}(v) = sign(v)\max(|v| - \lambda, 0) \quad (\text{III}.4)$$

where $\lambda$ is the threshold parameter, and $L_1$ refers to $Prox_{L_1}$ in the rest of the paper. Readers interested in the theory of proximal operations can refer to the book [21] therein for details. Eq. (III.3) is our critical idea, and the following two algorithms materialize the equation to generate adversarial examples.

*B. Integrated Gradient-based Finite Point Attack Algorithm (IFPA)*

As mentioned in Section II.A.3), we employ an integrated gradient method, which can reflect each input element's contribution to the logit output. Here, we apply the integrated gradient to select the perturbation points. It is intuitive that if we perturb the point that contributes the most to the results, the prediction result must change.

We follow this idea and the objective function described in Eq. (III.3) to design our Algorithm 1. In Algorithm 1, we can preset the number of perturbed points according to integrated gradients. That is, the bigger the value of integrated gradients, the more priority we have to perturbate the corresponding input points.

Concretely, by employing the integrated gradient algorithm,

we can obtain the integrated gradient matrix of $C^*(x)$ with respect to $x$. According to the input variable $\aleph$, we get the index positions of $\aleph$ maximum values in the gradient matrix via function $TopIndex(\cdot)$. Subsequently, $mask(\cdot)$ function generates a matrix only containing 0 and 1 based on the index positions, where the index positions are 1 and the rest are 0. The benefit of doing this is that we can perturb the points we want to, while the other points do not affect.

We carry out an iterative way to generate adversarial examples because BIM has proven that the iterative way is more efficient. Specifically, we define our objective function using logit output and $L_p$ distance discussed in Eq. (III.3), where $L_p$ can be $L_1$ or $L_2$. Then, Adam optimizer [31] is implemented to optimize the objective function, since it is proven the best choice among other optimizers [32]. The output of $Adam(\cdot)$ is the product of $\varepsilon$ and derivative matrix of objective function w.r.t. $x$, and the Hadamard product of $Adam(\cdot)$ and $mask(\cdot)$ is the perturbation we add to the legitimate $x$.

---

**Algorithm 1**

$x$ is the legitimate input, $C^*(x)$ is the label of $x$, *iters* is iterations, $Z(\cdot)$ is the function model learned, $\varepsilon$ is the perturbation rate, $\aleph$ is the number of perturbations, $c$ is the constraint parameter.

**Input:** $x, C^*(x), F, \varepsilon, \aleph$

1:    $x_{adv} = x$
2:    $ig_x = IntegratedGradient(x, Z(\cdot), C^*(x))$
3:    $max\_idx = TopIndex(ig_x, \aleph)$
4:    **for** $i=1$ **to** *iters*:
5:        $logits = Z(x_{adv})$
6:        $obj = logits[C^*(x)] + c \cdot L_p(x_{adv}, x)$
7:        $x_{adv} = x - Adam(obj, \varepsilon) \odot mask(max\_idx)$
8:    **end for**
9:    $x_{adv} = Clip(x_{adv}, 0, 1)$
10: **return** $x_{adv}$

---

**How do we combine $L_0$ with $L_1$ or $L_2$?** We believe that at the beginning, we preset the limitation of the number of perturbation points. Meanwhile, $L_0$ is the number of different points between $x$ and $x_{adv}$. Different points between $x$ and $x_{adv}$ is actually the embodiment of the perturbation points. Also, when generating adversarial examples, we use the $L_1$ or $L_2$ to restrain the perturbation as much as possible. So, we think this attack is a combination of $L_0$ and $L_1/L_2$. We call this attack integrated-based finite points attack (IFPA), and IFPA has variants according to the $L_p$. That is, when Eq. (III.3) uses $L_1$, we call IFPA-$R_1$ attack, and so on.

### C. *Integrated-based Universal Attack Algorithm (IUA)*

**In this section, we present** a new universal iterative attack. The pseudocode is shown in Algorithm 2. Algorithm 2 is similar to Algorithm 1 from the perspective of the objective function and the way to perturb the points. But the way to set the number of perturbation points is different. Specifically, it can be said that Algorithm 2 is an extension of Algorithm 1.

Concretely, in IFPA, we preset $\aleph$ before the algorithm 1 runs. But there are two potential problems: 1) $\aleph$ is not enough to generate an adversarial example; 2) $\aleph$ is too large when generating an adversarial example. To solve these, the number of perturbation points increases by 1 in each epoch from 1 perturbation point in Algorithm 2. This process is repeated until an adversarial example is generated, or we perturb all points. So, in this case, we need to get the perturbation order of all points, and that is what $sort(\cdot)$ function does. $sort(\cdot)$ function receives a multi-dimensional matrix and returns the index position of each value sorted from large to small. The rest part of Algorithm 2 is no different from Algorithm 1, and we refer to this algorithm as IUA. And we think this is another pattern to combine $L_0$ with $L_1/L_2$ since we do not perturb all the points at the beginning but perturb one by one. In this case, the perturbation will not be added to the impractical point to generate adversarial examples, and we call this algorithm **IUA**.

---

**Algorithm 2**

$x$ is the legitimate input, $C^*(x)$ is the label of $x$, *iters* is iterations, $Z(\cdot)$ is the function model learned, $\varepsilon$ is the perturbation rate, $c$ is the constraint parameter.

**Input:** $x, C^*(x), Z(\cdot), \varepsilon$

1:    $x_{adv} = x$
2:    $ig_x = IntegratedGradient(x, Z(\cdot), C^*(x))$
3:    $sorted\_idx = sort(ig_x)$
4:    $idx = 1$
5:    **while True:**
6:        **for** $i=1$ **to** *iters*:
7:            $logits = Z(x_{adv})$
8:            $obj = logits[y_{true}] + c \cdot L_p(x_{adv}, x)$
9:            $x_{adv} = x - Adam(obj, eps) \odot masked(sorted\_idx[:idx])$
10:       $x_{adv} = Clip(x_{adv}, 0, 1)$
11:       **if** $argmax(Z(x_{adv})) \neq C^*(x)$:
12:           **return** $x_{adv}$
13:       **if** $idx > len(sorted\_idx)$:
14:           **break**
15:    $idx \mathrel{+}= 1$
16:    **end for**
17: **end while**
18: **return** $x_{adv}$

---

### D. *Evaluation Metrics*

Four metrics will be used to evaluate our two algorithms. That is, crafting rate, the mean and standard deviation (std) of $L_0$, $L_1$ and $L_2$, where the definition of $L_p$ has been given in Section II.A.4). Here, $L_1$ is not the proximal operation but the formulation (II.2) when $p = 1$. The definition of crafting rate is shown below:

$$CraftingRate = \frac{\# \ of \ x_{adv}}{\# \ of \ x}$$

where # donates numbers. Crafting rate demonstrates the efficiency of the algorithm in generating adversarial examples. $L_0$ metric denotes how many points are perturbed. $L_1$ metric measures how many perturbations are added to the legitimate $x$. $L_2$ metric reflects Euclidean distance between $x_{adv}$ and $x$. These evaluation metrics comprehensively evaluate the quality of the generated adversarial examples. Mean and standard deviation of $L_p$ are used to quantify the distribution of perturbations under $L_p$ metric.

## IV. EVALUATION

This section first presents the datasets and models used in the experiments. Then, we discuss the results of IFPA under four metrics mentioned before. Also, analysis and comparison between IUA and other algorithms are described in Section IV.C.

TABLE II. THE STRUCTURE OF MNIST AND CIFAR-10 MODELS

| Layer | MNIST-model | CIFAR-10 model |
|---|---|---|
| ReLU(Convolution) | $3 \times 3 \times 32$ | $7 \times 7 \times 64$ |
| ReLU(Convolution) | $3 \times 3 \times 32$ | - |
| MaxPool | $2 \times 2$ | $3 \times 3$ |
| ReLU(Convolution) | $3 \times 3 \times 64$ | $\begin{bmatrix} 3 \times 3 \times 64 \\ 3 \times 3 \times 64 \end{bmatrix} \times 2$ |
| ReLU(Convolution) | $3 \times 3 \times 64$ | $\begin{bmatrix} 3 \times 3 \times 128 \\ 3 \times 3 \times 128 \end{bmatrix} \times 2$ |
| ReLU(Convolution) | - | $\begin{bmatrix} 3 \times 3 \times 256 \\ 3 \times 3 \times 256 \end{bmatrix} \times 2$ |
| ReLU(Convolution) | - | $\begin{bmatrix} 3 \times 3 \times 512 \\ 3 \times 3 \times 512 \end{bmatrix} \times 2$ |
| MaxPool | $2 \times 2$ | - |
| AvgPool | - | $7 \times 7$ |
| ReLU(FullyConnected) | 1024 | 1000 |
| Output | 10 | 10 |

### A. Description and Setup

In our experiment, we apply three datasets to do the experiments. The reason is that these three datasets, NSL-KDD [33], MNIST [34], CIFAR-10[35], are two types, where NSL-KDD is the structure dataset and the others are unstructured datasets. We want to cover as many different types of datasets as possible. The details of three datasets are shown in TABLE III, where the 'Size' column means the shape of one sample from the dataset, and 'Class' indicates the number of classes in each dataset.

Unfortunately, there is no baseline model to train the NSL-KDD dataset, so we have to stack a deep neural network to train it. The model structure is fc512-fc512-fc256-fc64-fc5, where fc512 means fully-connected layer with 512 nodes. Each layer uses *LeakyReLU* activation function. We use Adam optimizer to train the KDD-model. Other hyperparameters are as follows: learning rate is $10^{-4}$, epoch number is 20, batch size we set 32.

TABLE III. DATASET DESCRIPTION

| Dataset | Samples | | Size | Class |
|---|---|---|---|---|
| | Train | Test | | |
| NSL-KDD | 126003 | 22544 | 122 | 5 |
| MNIST | 60000 | 10000 | $1 \times 28 \times 28$ | 10 |
| CIFAR-10 | 50000 | 10000 | $3 \times 32 \times 32$ | 10 |

The performance of NSL-KDD model on the test set reaches an acceptable accuracy 78.96%.

In MNIST dataset, we train a MNIST-model described in TABLE II, where each convolution layer has a *ReLU* activation function. $3 \times 3 \times 32$ in the 1st line means this layer has 32 convolution kernels, each of whose shape is $3 \times 3$. We use SGD optimizer to train MNIST-model, where the learning rate we set 0.01, momentum is 0.5, the batch size is 32, and epoch is 10. Under these hyperparameters, the MNIST-model achieves 95.38% accuracy on the test set.

In CIFAR-10 dataset, we adopt the resnet-18 [36] to train the CIFAR10. Resnet is an epoch-making deep learning model. Its emergence solves the problem that the model is too deep to do backpropagation. The structure of resnet-18 is shown in TABLE II. $[3 \times 3 \times 64; \ 3 \times 3 \times 64] \times 2$ means that insiders of the matrix form a block, as shown in Fig. 4, and this block is stacked twice. Then, the block with different convolutional kernel sizes will be stacked together to form the resnet-18. We train the resnet-18 with SGD optimizer, whose learning rate is 0.1, momentum is 0.9. The epoch number and batch size we set are 300 and 128, respectively. Finally, resnet-18 achieves 95.36% accuracy on the test set.

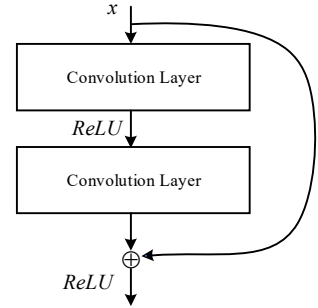

Fig. 4. The block forms the resnet-18.

One thing to declare is that all datasets need to preprocess before training the models. Unstructured datasets, MNIST and CIFAR10, are preprocessed by standardization:

$$x = \frac{x - x_{mean}}{x_{std}}$$

where $x_{mean}$ and $x_{std}$ are the mean and standard deviation of $x$.

The preprocessing of the structured NSL-KDD dataset is a little more complicated because there are continuous data and discrete data. We can use the function mentioned above to process the continuous data. As for discrete data, we implement the one-hot encoder to transform them. All models are trained with PyTorch library [37] and comparable algorithms described in Section II.C are implemented by PyTorch. We use an Intel Xeon E5-2650v4 CPU with a single NVIDIA GTX 1080Ti GPU for experiments on three datasets.

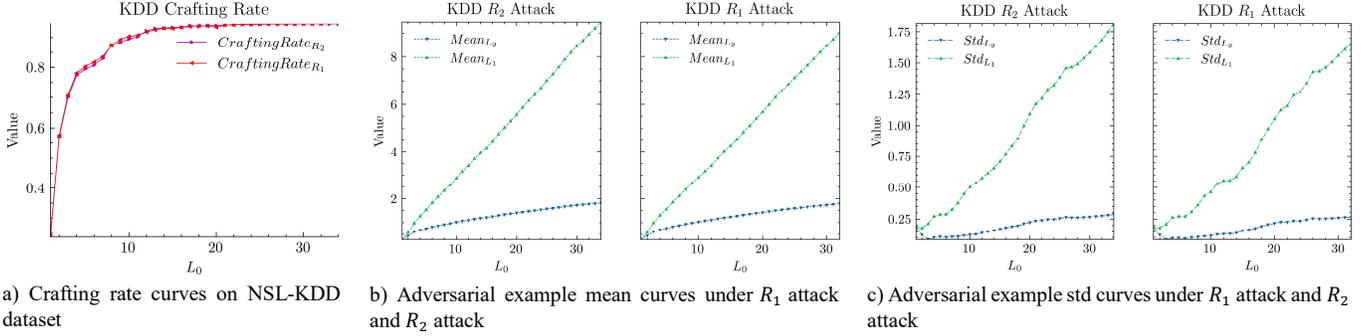

a) Crafting rate curves on NSL-KDD dataset  b) Adversarial example mean curves under $R_1$ attack and $R_2$ attack  c) Adversarial example std curves under $R_1$ attack and $R_2$ attack

Fig. 5. The performance of IFPA-$R_p$ attacks on NSL-KDD dataset under all metrics

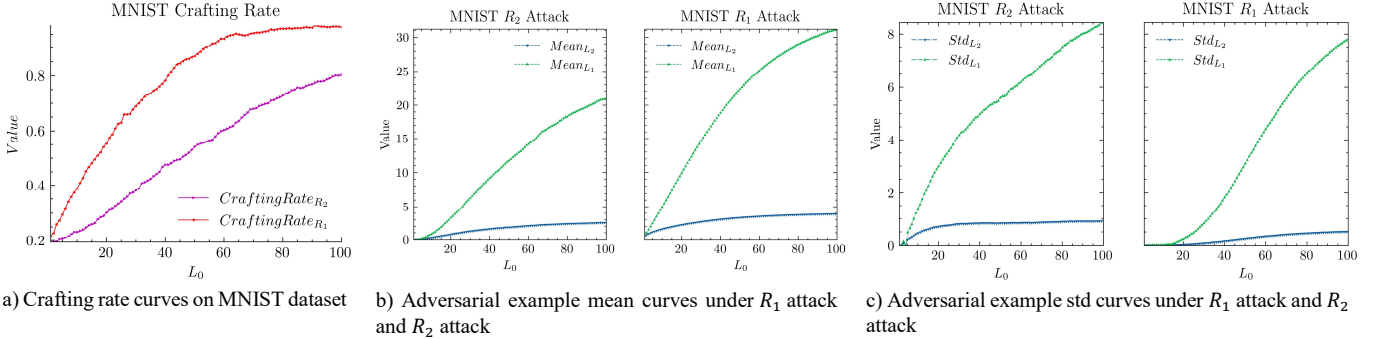

a) Crafting rate curves on MNIST dataset  b) Adversarial example mean curves under $R_1$ attack and $R_2$ attack  c) Adversarial example std curves under $R_1$ attack and $R_2$ attack

Fig. 6. The performance of IFPA-$R_p$ attacks on MNIST dataset under all metrics

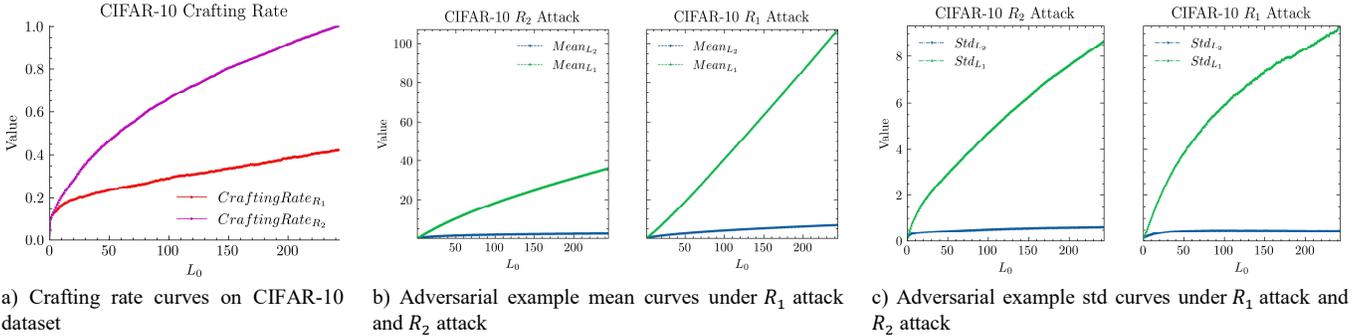

a) Crafting rate curves on CIFAR-10 dataset  b) Adversarial example mean curves under $R_1$ attack and $R_2$ attack  c) Adversarial example std curves under $R_1$ attack and $R_2$ attack

Fig. 7. The performance of IFPA-$R_p$ attacks on CIFAR-10 dataset under all metrics

## B. IFPA Results and Analysis

This section analyzes our proposed IFPA algorithm under the metrics described in Section III.D, as shown in Fig. 5, Fig. 6, Fig. 7. Considering the large number of repeated experiments, we randomly select 1000 samples in MNIST and CIFAR-10 datasets and 10000 samples in NSL-KDD dataset to evaluate our IFPA algorithms. The horizontal axis represents the number of perturbed points and we refer to points as $L_0$. The reason we have explained in Section III.B. Fig. 1 and Fig. 2 are adversarial examples generated by our IFPA on MNIST dataset and CIFAR-10 dataset, respectively.

Fig. 5.a displays the crafting rate curves under three datasets with $R_p$ using IFPA algorithm. As Fig. 5.a shows, there is no significant difference when using IFPA-$R_1$ attack and $R_2$ attack to generate adversarial examples on structured NSL-KDD dataset. Both curves show that the crafting rate increases gradually with perturbation points, which is intuitive. What a remarkable outcome is that ten perturbed points will produce a 90+% crafting rate.

We then analyze the perturbations of adversarial examples on NSL-KDD dataset. Fig. 5.b and Fig. 5.c provide the summary statistics for $L_p$ mean and standard deviation under IFPA $R_p$ attacks. We put the mean curves together for better comparison, as shown in Fig. 5.b. No significant differences between $Mean_{L_1}$ in $R_1$ and $R_2$ attacks are evident in Fig. 5.b. The same is true for $Mean_{L_2}$. As for std curves shown in Fig. 5.c, they are basically the same as that of mean curves, but the trend is not as smooth as the mean curves. Regardless of metrics of mean or std, the performance of $R_1$ attack is slightly better than $R_2$ attack. Overall, $R_1$ attack has an adversarial sample generation rate that is not inferior to $R_2$ attack, and the added perturbations are smaller than $R_2$ attack on NSL-KDD dataset.

As regards the crafting rate about MNIST dataset shown in Fig. 6.a, it is an interesting result that the crafting rate curve of IFPA-$R_1$ attack is higher than that of IFPA-$R_2$ attack, where 55 perturbed points will produce 90+% adversarial example generation rate under $R_2$ attack. The generation of $R_1$ attack is far less than $R_2$ attack. In Fig. 6.b and Fig. 6.c, however, no matter what metrics of $L_1$ and $L_2$, every element in $Mean_{L_p}$ of $R_1$ attack is higher than that of $R_2$ attack, which indicates that $R_2$ attack can not only constrain the $L_2$ distance between $x$ and $x_{adv}$, but also decrease the $L_1$ distance. One the contrary, $R_1$ attack cannot constrain the $L_1$ distance as well as $R_2$ attack. But $R_1$ attack is not without its merits. Asymmetrically, the standard deviation values of $R_1$ attack are lower than that of $R_2$ attack, where the $std_{L_2}$ curve of $R_1$ attack is significantly lower than that of $R_2$ attack, which indicate that the added perturbation of $R_2$ attack is not stable than $R_1$ attack. In short, $R_2$ attack on MNIST dataset has significantly smaller $mean_{L_p}$ values than $R_1$ attack. The drawbacks of $R_2$ attack is that the crafting rate is lower than $R_1$ attack and $Std_{L_p}$ values of $R_2$ attack are slightly higher than that of $R_1$ attack.

In terms of CIFAR-10 dataset, as shown in Fig. 7.a, the crafting rate curves of $R_p$ attacks are completely contrasting from those of $R_p$ attacks on MNIST. That is, the crafting rate curves of $R_2$ attack is much higher than that of $R_1$ attack. Especially, when the number of perturbation points is greater than 190, the crafting rate exceeds 90%. Fig. 7.b and Fig. 7.c show the performance of $R_p$ attacks under $L_p$ metrics. From the perspective of $Mean_{L_p}$ metrics, we can obtain the same conclusion discussed in Fig. 6. That is, $R_2$ attack can constrain both $L_1$ and $L_2$ distance between $x$ and $x_{adv}$, since $Mean_{L_p}$ values of $R_2$ attack are smaller than those of $R_1$. As Fig. 7.c shows, there is another significant difference between the two datasets. Concretely, $Std_{L_1}$ values of $R_2$ attack are smaller than that of $R_1$ attack on CIFAR-10 dataset, which differs from MNIST dataset. $Std_{L_2}$ values of $R_2$ attack are slightly larger than that of $R_1$ attack, which we think this can be negligible from a numerical perspective. In brief, $R_1$ attack is overall inferior to $R_2$ attack in CIFAR-10 data.

In summary, IFPA-$R_1$ attack achieves better results under $R_2$ attack on the structured NSL-KDD dataset. But on the unstructured dataset, the conclusion does not have uniformity. Specifically, $R_1$ attack is totally inferior to $R_2$ attack in CIFAR-10 dataset. But $R_1$ attack has more crafting rate than $R_2$ attack on MNIST dataset with more perturbations. We will explore the performance of IUA attack on these datasets to further compare $L_p$ in the next section.

### C. IUA Results and Analysis

In this section, we evaluate our proposed algorithms, integrated universal attack. As mentioned in Section II.C, we choose five adversarial generation algorithms to compare with our IUA algorithm. Here, we use $L_1$ and $L_2$ restriction to construct our IUA algorithms, respectively, and we denote them as IUA-$L_1$ and IUA-$L_2$.

TABLE IV provides the summary statistics of five algorithms and ours under all metrics on NSL-KDD structured dataset. From TABLE IV, we can see that each algorithm has little difference under the evaluation of Euclidean distance except PGD. Our IUA-$L_1$ algorithm achieve the second highest accuracy among them, where PGD is the most efficient to generate adversarial examples and its accuracy is 5% higher than ours. But under $L_p$ metrics, PGD performs worse than ours. Concretely, the number of perturbation points, $L_0$ mean value, of PGD is 60 times larger than that of IUA-$L_1$, which indicate that PGD needs more perturbation points to generate adversarial examples. The consequence of more perturbation points is that $L_1$ and $L_2$ values are very large, whatever the mean and std. The total perturbations, $L_1$ mean value of PGD is 15 times larger than that of IUA-$L_1$. It is worth thinking about the trade-off between perturbations and crafting rate. The values under metrics of IUA-$L_2$ are almost no differences with IUA-$L_1$. Other algorithms such as FGSM and BIM do not perform well than ours, but they confirm that the iterative way is better than one step to generate adversarial examples. DeepFool and CW, however, has the lowest crafting rate. This indicates that some adversarial algorithms are not universal in structured datasets and it is necessary to verify the structured dataset when proposing a new adversarial algorithm.

As for MNIST image unstructured datasets shown in TABLE V, our proposed IUA-$L_2$ achieves the best performance among comparable algorithms and IUA-$L_1$. Specifically, $L_p$ mean values are the smallest, indicating that the perturbation points and the total perturbations are the minima. Also, the highest crafting rate of IUA-$L_2$ also implies that our algorithm can generate high-quality adversarial examples with minimal perturbation and strong aggression. Although IUA-$L_1$ attack acquires the second-highest crafting rate. The perturbations are also added relatively more, which we conjecture $L_1$ restriction is not as suitable as $L_2$ restriction for unstructured datasets.

Other algorithms, such as BIM and PGD, perform well since their crafting rates reach 90+%. The perturbation of PGD are slightly larger than BIM, but these perturbations of the two algorithms are much larger than that of our IUA-$L_2$. Concretely, the number of perturbation points and total perturbations are nine times and four times larger than ours, respectively. FGSM, DeepFool and CW algorithms perform badly. All of them have almost the same $L_p$ values as PGD, but their crafting rates are lower. We can draw a conclusion from 0 clearly: our proposed IUA does restrict the perturbation points and total perturbations at the same time while not decreasing the crafting rate.

The results under all metrics of CIFAR-10 dataset is shown in TABLE VI. In this dataset, we can see that our proposed IUA-$L_2$ achieves the lowest perturbation points and the lowest perturbations though, its crafting rate is lower than CW and PGD. Considering the lowest perturbation points and perturbations, the crafting rate of 83.24% is tolerable. Specifically, in terms of $L_0$ mean values, all 5 comparable algorithms are ten times larger than our IUA-$L_2$. Although high $L_0$ std value implies the perturbation points of adversarial examples generated by IUA-$L_2$ fluctuate in a wide range, other algorithms are still larger than ours according to the three-sigma rule. In respect of $L_2$ metrics, our mean value also achieves the best and the std gets the second-smallest value. But the performance of IUA-$L_1$ proves that $L_1$ restriction is not suitable for unstructured datasets and $L_2$ restriction can better restrict the perturbation of unstructured datasets.

TABLE IV. COMPARISON BETWEEN FIVE ALGORITHMS AND OUR IUA ALGORITHM ON NSL-KDD DATASET

| NSL-KDD | Crafting Rate | $L_0$ | | $L_1$ | | $L_2$ | |
|---|---|---|---|---|---|---|---|
| | | Mean | Std | Mean | Std | Mean | Std |
| IUA-$L_1$ | *93.43%* | **4.63** | *8.68* | 1.25 | 2.25 | 0.50 | 0.48 |
| IUA-$L_2$ | *93.38%* | *4.84* | 9.88 | 1.24 | 2.15 | 0.49 | 0.47 |
| FGSM | 86.67% | 60.75 | 12.55 | 5.89 | 1.26 | 0.76 | 0.08 |
| BIM | 89.55% | 75.16 | 15.83 | 5.24 | 0.96 | 0.66 | **0.07** |
| DeepFool | 44.00% | 58.16 | 11.28 | **0.31** | **0.39** | **0.08** | 0.08 |
| CW | 90.10% | 121.78 | **1.33** | 0.63 | 0.66 | 0.16 | 0.16 |
| PGD | **98.62%** | 120.02 | 14.45 | 19.29 | 1.94 | 3.02 | 0.23 |

TABLE V. COMPARISON BETWEEN FIVE ALGORITHMS AND OUR IUA ALGORITHM ON MNIST DATASET

| MNIST | Crafting Rate | $L_0$ | | $L_1$ | | $L_2$ | |
|---|---|---|---|---|---|---|---|
| | | Mean | Std | Mean | Std | Mean | Std |
| IUA-$L_1$ | 96.68% | 333.26 | 328.63 | 678.52 | 699.03 | 31.31 | 20.91 |
| IUA-$L_2$ | **99.27%** | **82.71** | 38.58 | **129.70** | *62.25* | **14.60** | 4.20 |
| FGSM | 63.02% | 746.26 | 9.47 | 597.01 | 7.57 | 21.85 | 0.14 |
| BIM | 91.00% | 783.38 | 0.81 | 509.64 | 47.13 | 21.81 | 2.52 |
| DeepFool | 54.00% | 784.00 | 0.00 | 433.77 | 43.43 | 19.07 | 2.50 |
| CW | 99.22% | 773.99 | 0.16 | 446.21 | 46.03 | 19.83 | 2.59 |
| PGD | 93.90% | 784.00 | 0.16 | 546.34 | 77.99 | 27.95 | 4.26 |

TABLE VI. COMPARISON BETWEEN FIVE ALGORITHMS AND OUR IUA ALGORITHM ON CIFAR-10 DATASET

| CIFAR10 | Crafting Rate | $L_0$ | | $L_1$ | | $L_2$ | |
|---|---|---|---|---|---|---|---|
| | | Mean | Std | Mean | Std | Mean | Std |
| IUA-$L_1$ | 78.00% | 347.74 | 509.38 | 1042.01 | 359.75 | 41.94 | 38.35 |
| IUA-$L_2$ | 83.24% | **283.46** | 471.62 | **235.52** | **267.90** | **26.76** | *14.40* |
| FGSM | 58.25% | 3072.00 | 0.00 | 2214.83 | 857.92 | 50.74 | 15.82 |
| BIM | 84.08% | 2990.83 | 59.60 | 1603.94 | 568.64 | 35.97 | 10.21 |
| DeepFool | 52.63% | 3072.00 | 0.06 | 2188.91 | 885.60 | 51.20 | 15.88 |
| CW | **91.59%** | 3055.07 | 22.55 | 2121.59 | 882.18 | 50.19 | 15.81 |
| PGD | 89.87% | 3072.00 | 0.00 | 3186.35 | 962.70 | 66.97 | 17.78 |

In all datasets, BIM and PGD achieve good results in terms of crafting rates, but no restriction terms lead to great perturbations. FGSM, DeepFool, and CW did not perform as well as we thought, especially in structured datasets. This strongly indicates that it is critical to verify the proposed algorithm on all types of datasets. Meanwhile, our proposed IUA algorithm is able to handle all type datasets and restrict the perturbations based on $L_p$ restriction. Concretely, $L_1$ restriction performs well on structured datasets but performs badly on unstructured datasets. $L_2$ restriction is suitable for unstructured dataset than $L_1$.

*D. Summary*

We make a summary of the extensive results and analysis from Section IV.B and Section IV.C. No matter what IFPA and IUA, $L_1$ restriction have achieved better results than $L_2$ slightly on the structured dataset. However, $L_2$ restriction is much better than $L_1$ on unstructured datasets, and $L_2$ restriction can limit both $L_2$ distance and $L_1$ distance.

V. CONCLUSION AND FUTURE WORK

In this paper, we propose two integrated-based adversarial generation algorithms, IFPA and IUA, which can be used to conduct white-box attacks to DNNs. The extensive experimental results verify that our algorithms can generate high-quality adversarial examples on all types of datasets with lower perturbations and higher crafting rates. Meanwhile, the results indicate that $L_2$ restriction is more suitable for unstructured datasets than $L_1$, and $L_1$ restriction is better than $L_2$ on structured datasets.

In future work, a targeted adversarial attack based on the integrated gradient will be considered. In addition, the interpretation of why $L_2$ restriction performs well on unstructured datasets should be explored from the perspective of mathematical theory.